\documentclass[10pt]{article} %
\usepackage[preprint]{rlc}

\usepackage{amssymb}            %
\usepackage{mathtools}          %
\usepackage{mathrsfs}           %
\mathtoolsset{showonlyrefs}     %
\usepackage{graphicx}           %
\usepackage{subcaption}         %
\usepackage[space]{grffile}     %
\usepackage{url}                %

\usepackage{amsmath}
\usepackage{amssymb}
\usepackage{mathtools}
\usepackage{amsthm}
\usepackage{amsfonts}
\DeclareMathOperator*{\argmax}{argmax}
\usepackage{multirow}

\theoremstyle{plain}
\newtheorem{theorem}{Theorem}[section]

\theoremstyle{definition}
\newtheorem{definition}[theorem]{Definition}

\theoremstyle{remark}

\title{Tiered Reward: Designing Rewards for \\Specification and Fast Learning of Desired Behavior}

\author{Zhiyuan Zhou
    \thanks{Correspondence to: zhiyuan\_zhou@berkeley.edu. Code for the paper can be found at \url{https://github.com/zhouzypaul/tiered-reward}}\\
    UC Berkeley
    \And 
    Shreyas Sundara Raman \\
    Brown University 
    \And 
    Henry Sowerby \\
    Brown University 
    \And 
    Michael L. Littman \\
    Brown University
}

\begin{document}

\maketitle

\begin{abstract}
Reinforcement-learning agents seek to maximize a reward signal through environmental interactions. As humans, our job in the learning process is to design reward functions to express desired behavior and enable the agent to learn such behavior swiftly. However, designing good reward functions to induce the desired behavior is generally hard, let alone the question of which rewards make learning fast. In this work, we introduce a family of a reward structures we call Tiered Reward that addresses both of these questions. We consider the reward-design problem in tasks formulated as reaching desirable states and avoiding undesirable states. To start, we propose a strict partial ordering of the policy space to resolve trade-offs in behavior preference. We prefer policies that reach the good states faster and with higher probability while avoiding the bad states longer. Next, we introduce Tiered Reward, a class of environment-independent reward functions and show it is guaranteed to induce policies that are Pareto-optimal according to our preference relation. Finally, we demonstrate that Tiered Reward leads to fast learning with multiple tabular and deep reinforcement-learning algorithms.
\end{abstract}

\section{Introduction}
\label{sec:intro}

Reinforcement learning~\citep{sutton98} (RL) is concerned with the problem of learning to behave to maximize a reward signal.
In biological systems, this reward signal is considered to be the organism's motivational system, using pain and pleasure to modulate behavior~\citep{porreca2017reward, leknes2010pain, navratilova2014reward}.
In engineered systems, however, rewards must be selected by the system designer~\citep{nagpal2020reward, dewey2014reinforcement}. 
We view rewards as a kind of programming language---a specification of the agent's target behavior~\citep{gltl, zhou2022programmatic}. 
As arbiters of behavior correctness in the learning process, humans bear the responsibility of authoring this program.
This is referred to as the reward-design problem~\citep{dewey2014reinforcement, devidze2021explicable, sorg2010reward, sowerby2022designing}: \textit{given a set of desired behavior, what kind of reward functions would efficiently express these behavior?}
In this paper, we look at rewards that can correctly and efficiently express desirable states (goals and subgoals) and undesirable states (obstacles).

There are two essential steps in designing reward functions. First, one must decide what kind of behavior is desirable and should be conveyed. Then, there's the choice of reward function that induces such behavior.
The first step is hard because there is no universal preference over behavior and having to explicitly write down all possible trade-offs is challenging.
Even if the reward designer has a way of expressing preferences for all possible exchanges, it can be difficult, or even impossible, to design a reward function that captures them without prior knowledge of the environment.

First, let us consider the question of how to specify preference over a set of policies~\citep{roy2021direct}. 
In the goal--obstacle class of tasks we consider, preferences over policies are simple in deterministic environments.
We imagine all states are either goal states, obstacle states, or neither (background states), and all goals and obstacles are absorbing. 
Preferences in deterministic environments form a total order: reaching goals is better than reaching obstacles; if the policy reaches goals, faster is better; if the policy reaches obstacles, avoiding them longer is better.
However, even in this simple setting, providing such precise trade-offs is difficult in stochastic environments.
Is it better for an agent to increase the chance of getting to the goal by 5\% if it also incurs 8\% higher probability of hitting an obstacle? Is it better to increase by 50\% the probability of getting to a goal if the expected time of getting there also increases by 20\%? 
Preferences are less clear in a stochastic setting because there can be trade-offs between different outcomes and their probabilities. 
However, some comparisons are arguably clear cut. 
Informally, if one policy induces uniformly better outcomes than another---being more likely to reach a goal and doing so faster, being less likely to reach an obstacle and getting there more slowly---we prefer such a policy. 
If the policies can't be directly compared, we propose to be indifferent between them. Thus, we replace the standard reinforcement-learning notion of optimality with Pareto-optimality~\citep{mornati2013pareto} which is commonly adopted in multiobjective RL~\citep{vamplew11, van2014multi}; we seek a policy that is either preferred or incomparable to every other policy. 

Even after resolving the issue of specifying behavior preference, policies are hard to express through reward functions in general~\citep{zhang2009policy, amodei2016faulty}, and some are even impossible to convey with a Markov (state--action-based) reward~\citep{abel2021expressivity}. 
Even when policies are expressible, designing bad reward functions can lead to undesirable or dangerous actions~\citep{amodei2016faulty}, easy reward hacking~\citep{amodei2016concrete, skalse2022defining}, and more. We seek to design good reward functions, which can be characterized by many properties, such as interpretability and learning speed~\citep{devidze2021explicable}. But the most important property a reward function must have is to guarantee the adoption of a desired policy. As we will show later in Section~\ref{sec:tiered-reward}, even intuitively correct reward designs can lead to suboptimal policies. To hedge against this, we introduce a tiered reward structure that is guaranteed to induce Pareto-optimal policies. 
Intuitively, we partition the state space into several tiers, or goodness levels. States in the same tier are associated with the same reward, while states in a more desirable tier are associated with an exponentially higher reward. We prove that these tiered reward structures, with the proper constraints between reward values, induce Pareto-optimal behavior and empirically show that they can lead to fast learning. 

In this paper, we propose Tiered Reward as a way to design reward functions that can correctly and swiftly express desired behavior.
Our contribution is threefold: First, we define a preference over the entire policy space via a strict partial ordering on outcomes using the notion of Pareto-optimality.
This addresses the question of behavior preference in stochastic environments.
Then, we introduce a class of environment-independent tiered reward structure that provably induce Pareto-optimal policies with respect to this preference ordering. 
Finally, we demonstrate these tiered reward functions can lead to fast learning in both tabular and deep RL settings and is invariant to the choice of RL algorithms.

\section{Related Work}

\paragraph{Specifying behavior through rewards: }
Preference-based RL methods~\citep{wirth2017survey, brown2019extrapolating, liu2022meta} learn a reward function based on a dataset of preferences over trajectories. But, as we have argued, preferences over trajectory probabilities can be very difficult to specify. In addition, our reward scheme relieves the need for environment-specific preference datasets created by human experts. 
Multi-objective RL~\citep{vamplew11, toro2018teaching, hayes2022practical} allows for different tasks to be specified through a set of reward functions. Our work proceeds in the orthogonal direction by designing a single reward function to trade off among multiple behaviors, instead of incentivizing all of them.
Reward machines~\citep{icarte2018using, icarte2022reward} are finite state machines that compose reward functions and allow different rewards to be delivered dependent on the agent's trajectory. They reveal the structure of the reward function to the RL agent to support decomposition of complex tasks. Our focus on how to provide incentives for specific outcomes is complementary and the two approaches can be used in concert. 
Temporal logic based languages~\citep{gltl, camacho2017non, li2017reinforcement, camacho2019ltl} have been used to specify behavior. Though these methods can be more expressive, they often lead to intractable planning and learning problems due to state-space explosion issues~\citep{wongpiromsarn2010receding}. We offer a different expressibility--tractability tradeoff. 

\paragraph{Reward Design for fast learning: }
\citet{mataric1994reward} proposed to accelerate learning by incorporating domain knowledge and using a progress estimator, but does not provide a principled method of designing a reward function. 
\citet{sowerby2022designing} showed that reward functions that maximize the action gap given a measure of horizon length lead to fast learning. 
However, designing such reward functions requires solving an optimization problem with detailed knowledge of the environment, making this approach impractical.
Similarly, \citet{devidze2021explicable} formulated the reward-design problem as an optimization problem to maximize informativeness and sparsity.
However, their method requires solving the MDP with ground truth transition dynamics and a reference reward function, which is often not available in practice.

\section{Problem Setting}
\label{sec:background}

We view an RL environment as a Markov Decision Process (MDP), with state space $S$, action space $A$, transition model $T$, reward function $R$, and discount factor $\gamma$. A policy $\pi: S \times A \rightarrow [0, 1]$ is a mapping from the current state to a probability distribution of the action to be taken. 
The optimal policy starting from some initial state $s_0$ in the MDP is defined as any reward-maximizing policy $\pi^* \in  \argmax_\pi \mathbb{E}[\sum_t \gamma^t r_t | s_0, \pi]$. 
To make the reward-design problem as simple as possible for designers, we limit the reward function $R: S \rightarrow \mathbb{R}$ to be defined solely on states. In goal--obstacle tasks, we consider the goal states and obstacle states to be absorbing.

We are interested in the reward-design problem. In the common RL framework, tasks are specified by the reward function and the agent’s objective is to maximize cumulative rewards. We take an alternative perspective: We specify tasks by prescribing a set of desirable policies and seek reward functions such that maximizing the reward will lead to the desirable policies. Formally, given a set of desirable (Pareto-optimal) policies $\Pi$, the reward-design problem is to create a reward function $R : S \rightarrow R$ such that the optimal policy $\pi_R^* \in \Pi$.

We imagine the state space $S$ as exhibiting a tiered structure, where higher tiers are more desirable than lower tiers, and states within the same tier are equally desirable. Formally, we define:

\newcommand{\cupdot}{\mathbin{\mathaccent\cdot\cup}}

\begin{definition}
    $k$-Tier Markov Decision Process: A Tier MDP is an MDP with state space $S$, action space $A$, transition model $T: S \times A \times S \rightarrow \mathbb{R}$, reward function $R: S \rightarrow \mathbb{R}$, and discount factor $\gamma$. The state space is partitioned into $k$ tiers, where $S = S_1 \cupdot S_2 \cupdot \ldots \cupdot S_k$ and $S_i \cap S_j = \emptyset, \forall i \neq j \in 1, 2, ..., k$. The reward function has the form $R(s) = r_i, \forall s \in S_i, i=1,2,...,k$. In addition, $r_1 < r_2 < \ldots < r_k$.
\end{definition}

\begin{figure}[htbp]
    \centering
    \hfill
    \raisebox{-0.5\height}{%
        \begin{subfigure}[b]{0.4\textwidth}
            \centering
            \includegraphics[width=0.5\textwidth]{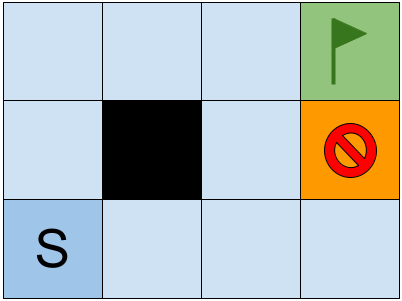}
            \caption{Russell/Norvig grid world. Objective is to reach the goal (green) without first visiting lava (orange) ($\gamma=0.9$).}
            \label{fig:rn-grid}
        \end{subfigure}%
    }
    \hfill%
    \raisebox{-0.5\height}{%
        \begin{subfigure}[b]{0.5\textwidth}
            \centering
            \includegraphics[width=0.5\textwidth]{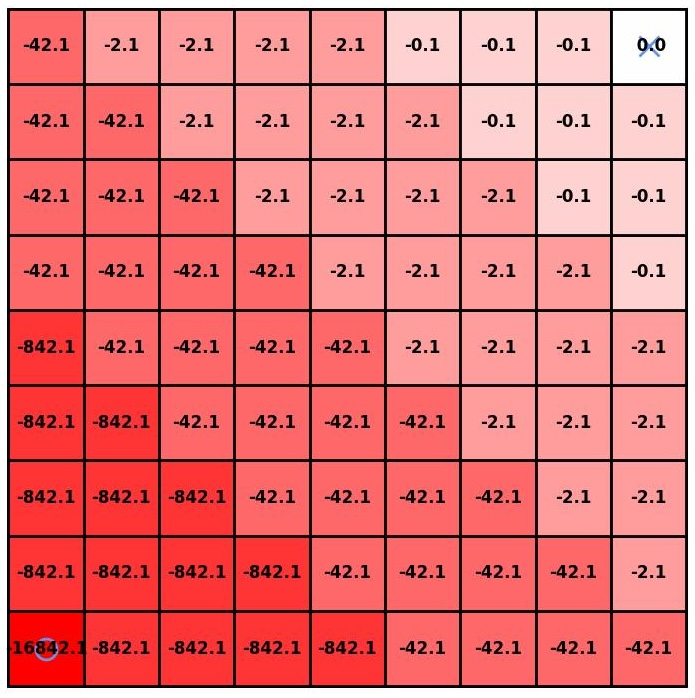}
            \caption{A Tiered Reward in a grid world with $6$ tiers. Start state in bottom left, goal state in top right. Darker colors correspond to more negative reward values.}
            \label{fig:grid-tier}
        \end{subfigure}%
    }
    \hfill%
    \caption{}
\end{figure}

As an example, the grid world from~\citet{russell2010artificial}, as illustrated in Figure~\ref{fig:rn-grid}, could be formulated as a 3-Tier MDP---the goal state is one tier ($S_3$), the lava state one tier ($S_1$), and all other states reside in the background tier ($S_2$). It is important to note that we put no constraints on how many states could be in each tier, nor how many tiers there can be. Therefore, the framework has a high degree of generality: any finite MDP with reward defined on states could be formulated as a Tier MDP by placing states with the same reward in the same tier. However, the Tier MDP is most useful when there are clear good and bad states in the state space, such as when there are goal and obstacle states, or even states of intermediate desirability such as subgoal states. In the following sections, we will show how to perform reward design in Tier MDPs.

\section{Policy Ordering with Pareto-optimality}
\label{sec:pareto}
A policy induces a probability distribution over an infinite set of outcomes (specifically the probability of reaching each of the states after $t$ steps, for all $t$). In goal--obstacle tasks, policies can be characterized by statistics such as probability of reaching the goal and probability of avoiding the obstacle for each possible horizon length. Using these statistics, we will show below how the entire policy space can form a strict partial ordering to specify which policies are preferable.

We define $o_t$ to be the probability of being in $S_1$ at timestep $t$, and $g_t$ that of $S_3$. Given two policies $\pi^A$ and $\pi^B$, we say $\pi^A$ \emph{dominates} $\pi^B$ when both of these inequalities hold (and not both being strictly equal at all times): 
$$\sum_{i=0}^{t} o_i^A \leq \sum_{i=0}^{t} o_i^B, \quad %
\sum_{i=0}^{t} g_i^A \geq \sum_{i=0}^{t} g_i^B, \quad \quad \forall t = 0,1,2,...,\infty.$$
In words, one policy dominates another if it gets to the goal faster, while delaying encountering obstacles longer. The set of policies that are not dominated by any other policy is the set of \emph{Pareto-optimal} policies. 
Because there is a finite number of policies and domination is transitive, the set of Pareto-optimal policies is non-empty. See a visualization of Pareto-optimal policies in Appendix~\ref{apdx:pareto}.

Pareto-optimal policies are interesting to consider for two main reasons. First, Pareto-optimal behavior always exists, even when policies that achieve other reasonable things do not. 
Secondly, Pareto-optimality addresses the preference problem by defining a strict partial ordering over the entire policy space. Although the policies on the Pareto frontier are incomparable among themselves, they are all better than the set of Pareto-dominated policies. We simply deem the set of Pareto-optimal policies to be the desirable behavior, and all others undesirable. Next, we show how to design rewards that guarantee Pareto-optimal policies.

\section{Tiered Reward}
\label{sec:tiered-reward}

In this section, we seek a sufficient condition on the reward function so that optimizing expected discounted reward will result in a Pareto-optimal policy with respect to our preference relation.
\begin{definition}
    Pareto-optimal rewards: A reward function $R(s)$ is called \emph{Pareto-optimal} if the policy it induces, $\pi_R \in \argmax_\pi \mathbb{E}[\sum_t \gamma^t r_t | s_0, \pi]$, is Pareto-optimal.
\end{definition}

Even some reasonable-sounding reward functions need not be Pareto-optimal. Going back to the Russell/Norvig grid example, an intuitive reward design would be requiring $r_{lava} < r_{background} < r_{goal}$. Consider three example reward functions in Table~\ref{tab:three-reward-functions} that satisfy this constraint:
Both $R$ and $G$ are Pareto-optimal, while $B$ is Pareto-dominated (see Figure~\ref{fig:rgb-policy} in Appendix). Roughly, $B$ doesn't encourage getting to the goal but is also not good at avoiding lava.

\begin{table}[htb]
    \centering
    \begin{tabular}{|c|c|c|c|}
    \hline
    Policy &  $r_{lava}$ & $r_{background}$ & $r_{goal}$ \\
    \hline
    R & $-1$ & $-0.1$ & $+1$ \\
    \hline
    G & $-1$ & $0$ & $+0.5$ \\
    \hline
    B & $-1$ & $-0.9$ & $0$ \\
    \hline
    \end{tabular}
    \caption{Three intuitive reward functions of Russell/Norvig grid world. $B$ is Pareto-dominated.}
    \label{tab:three-reward-functions}
\end{table}

In fact, many of the reward functions that satisfy $r_{lava} < r_{background} < r_{goal}$ are not Pareto-optimal. Out of $1000$ such rewards that we sampled randomly, 90.5\% were Pareto-dominated (See Figure~\ref{fig:pareto-random-rewards} in Appendix). Next, we present a simple rule that is sufficient to guarantee environment-independent Pareto-optimal reward functions.

\subsection{The $3$ Tiers Case}
To facilitate understanding, we limit the problem space to 3-Tier MDPs for now, and generalize to $k$-Tier MDPs in Section~\ref{sec:k-tiers}. In a 3-Tier MDP, we will call the 3 tiers obstacles ($S_1$), background ($S_2$), and goals ($S_3$), in order of increasing desirability. States in $S_1$ and $S_3$ are absorbing.  
 
\begin{definition}
\label{def:3-tier-reward}
    Tiered Reward: In a $3$-Tier Markov Decision Process with discount factor $\gamma \in (0, 1)$, a reward function defined by
    \begin{equation*}
        R(s) = 
            \begin{cases}
            r_{obs} & \text{if\ } s \in S_1 \\
            r_{background} & \text{if\ } s \in S_2 \\
            r_{goal} & \text{if\ } s \in S_3 \\
            \end{cases}
    \end{equation*}
    is considered a \emph{Tiered Reward} if 
    $$r_{obs} < \frac{1}{1-\gamma} r_{background} < r_{goal}.$$
    and states in $S_1$ and $S_3$ are absorbing.
\end{definition}

\begin{theorem}[Pareto-optimal rewards in 3-Tier MDP]
\label{thm:three-tier}
In a 3-Tier Markov Decision Process, a Tiered Reward is Pareto-optimal. 
\end{theorem}

We leave the proof in Appendix~\ref{apdx:thm-1-proof} but provide some intuition for Tiered Reward here. The middle term in Definition~\ref{def:3-tier-reward}, $\frac{1}{1-\gamma} r_{background}$, is equal to the cumulative discounted return for infinitely getting a reward in the background tier ($(1 + \gamma + \gamma^2 + ...)r_{background}$). So, in a gross simplification, as long as the reward at the goal is more appealing than infinitely wandering in background states, and the obstacle less appealing, the reward induces behavior that arrives at the goal early and avoids the obstacles.
Following this simple constraint, we as reward designers can easily create Pareto-optimal reward functions without requiring knowledge of the transition probabilities in the environment. 
Though environment-specific knowledge is needed partition the state space into tiers, Tiered Reward is generally applicable and environment-independent in the sense that the reward structure remain the same and the reward values for each tier are shared across different MDPs.

\subsection{Generalizing to $k$ Tiers}
\label{sec:k-tiers}

MDPs with more than $3$ tiers can usefully model important problems such as those with well-defined subgoal states. Specifically, each subgoal region could be its own tier, instead of being grouped into one big background tier. Even though these problems could still be solved as a $3$-Tier MDP, more knowledge about the environment could help design better reward functions and accelerate learning. 

\begin{definition}
\label{def:k-tier-reward}
    Tiered Reward: In a $k$-Tier($k > 3$) Markov Decision Process with discount factor $\gamma \in (0, 1)$ where the goal tier ($k$) is absorbing, the reward function $R$ is a Tiered Reward if $R(s) = r_i, \forall s \in S_i, i=1,2,...,k$, for reward values $r_1, r_2, ..., r_k \in \mathbb{R}$, that satisfy
    $$r_1 < (\frac{1}{1-\gamma}) r_2 < (\frac{1}{1-\gamma})^2 r_3 < \cdot \cdot \cdot < (\frac{1}{1-\gamma})^{k-1} r_k \leq 0.$$
\end{definition}

One such reward is visualized in Figure~\ref{fig:grid-tier}.
Notice that the $k$-Tiered Reward (Definition~\ref{def:k-tier-reward}) uses a stricter condition than that of $3$-Tiered Reward (Definition~\ref{def:3-tier-reward}).
In fact, Definition~\ref{def:k-tier-reward} is a sufficient condition for Definition~\ref{def:3-tier-reward} because $3$ tiers is a special case with only one non-absorbing tier. One can also design a stricter $3$-Tiered Reward with Definition~\ref{def:k-tier-reward} with $k=3$.
Definition~\ref{def:k-tier-reward} is stricter in that:
first, all reward values are non-positive. This is a sufficient but not necessary condition to guarantee Pareto-optimality. We enforce this constraint not only to make it a sufficient condition, but also because step-wise penalty has been proved to support faster learning~\citep{koenig1993complexity, sowerby2022designing}. Specifically, with a zero-initialized value function, step penalties create an incentive for the agent to try state--action pairs it has never experienced before, resulting in rapid exploration. Secondly, the reward values of higher tiers are exponentially greater than the lower ones. For adjacent tiers $i$ and $i+1$, The reward values always satisfy $r_i < \frac{1}{1-\gamma} r_{i+1} < 0$.

This definition can be understood as a generalization of the 3-Tiered Reward. When the agent resides within tier $i \in \{2, 3, ..., k-1\}$, the $k$ tiers could be partitioned into $3$ groups to construct a 3-Tier MDP. In particular, $S_1$ will include tiers $1$ through $i-1$, $S_2$ is just tier $i$, and $S_3$ is tiers $i+1$ to $k$. Note that we can generalize Theorem~\ref{thm:three-tier} to allow states in $S_1$ and $S_3$ to have any reward values as long as they satisfy the inequality in Definition~\ref{def:3-tier-reward} for a fixed reward value in $S_2$. Namely, denote $r_{low} = \max \{r_1, ...,r_{i-1}\}$ and $r_{high} = \min \{r_{i+1}, ..., r_k\}$, and as a $k$-Tiered Reward they satisfy
 $r_{low} < (\frac{1}{1-\gamma}) r_i < (\frac{1}{1-\gamma})^2 r_{high}$
And since $\gamma \in (0, 1)$, 
$    r_{low} < (\frac{1}{1-\gamma}) r_i < (\frac{1}{1-\gamma})^2 r_{high} \leq r_{high}. $
That is, $(r_{low}, r_i, r_{high})$ is a Tiered Reward function in the $3$-Tier MDP with tiers $S_1$, $S_2$, and $S_3$, and  therefore induces Pareto-optimal policies (Theorem~\ref{thm:three-tier}). So, at tier $i$, the policy that optimizes the $k$-Tiered Reward will push agents to higher tiers as fast as possible and avoid lower tiers, as if they were goals and obstacles, respectively. In the special case that the agent resides within tier $i=1$, the constraint from Definition~\ref{def:k-tier-reward} will treat tiers $2$ through $k$ as if they are all goals, pushing the agent towards them. In the case that $i=k$, the agent is already in the ``goal tier". So overall, $k$-Tiered Reward will induce in a ratchet-like policy---go to the higher tiers as fast as possible while not fall back to the lower tiers---that makes learning fast. In fact, it has been shown that a similar increasing reward profile~\citep{sowerby2022designing} leads to fast learning. \citet{okudo2021subgoal} and \citet{zhai2022computational} have also shown that intermediate rewards can accelerate learning and provably improve sample efficiency in goal--reaching tasks.

Besides encouraging early visitation of good tiers, using Tiered Reward also guarantees maximum total visitation of all good tiers. This property is formalized in Theorem~\ref{thm:multi-tier}.

\begin{theorem}[Tiered Reward and Cumulative Tier Visitation]
\label{thm:multi-tier}
In a $k$-Tier Markov Decision Process that has Tiered Reward $R(s)$, the induced optimal policy is $\pi^*$. Let ${p_t^*}^d \in [0, 1]$ be the probability of being in tier $d \in \{1,2, ..., k\}$ for the first time at timestep $t$ following policy $\pi^*$. Then, there is no policy $\pi$, along with its induced probability distribution $p_t^d$, that satisfies both:
$$ \sum_{i=0}^{t} p_i^1 \leq \sum_{i=0}^{t} {p_i^*}^1, \quad \forall t = 0, 1,2,...,\infty  \quad \quad%
 \sum_{i=0}^{t} p_i^d \geq \sum_{i=0}^{t} {p_i^*}^d, \quad \forall d = [2..k], \forall t = 0, 1,2,...,\infty.$$
\end{theorem}
The proof is similar to that of Theorem~\ref{thm:three-tier} and can be found in Appendix~\ref{apdx:thm-multi-tier-proof}.
To state the theorem in words, if a $k$-Tier MDP has a Tiered Reward structure, then the resulting policy will visit the worst tier ($S_1$) for as few times as possible, while visiting all the other good tiers ($S_2$, ..., $S_k$) as often as possible, respectively.

\section{Experiments} 
\label{sec:experiment}
Guaranteeing Pareto-optimal behavior is not the sole benefit of using Tiered Rewards; we find that it also leads to fast learning compared to two baseline reward functions.
The first one, following \citet{koenig1993complexity}, we call \emph{action penalty}. This reward penalizes each step with $-1$, until the goal state is reached and the agent is awarded $+1$. To reiterate, such step-wise negative reinforcement encourages directed exploration assuming only knowledge about position of the goal.
The second one uses reward shaping as ~\citet{ng1999policy} showed that subgoals can be leveraged through shaping rewards to guide the learning process.
Reward shaping using an optimal value function is not a fair comparison both because it contains more information about the environment than tiers and it requires solving the environment with a pre-specified ground truth reward function, which induces a circular logic in reward design.
For direct comparability, we use the Tiered Reward as a potential function $\Phi(s)=R_{tier}(s)$ to shape the action-penalty reward, resulting in what we call \emph{tier-based shaping reward}: $R_{tbs}(s, a, s')= R_{penalty}(s) + \gamma \Phi(s') - \Phi(s)$. 

There are many ways to design a Tiered Reward because it is a class of reward functions that is only constrained by an inequality (Definition~\ref{def:k-tier-reward}), and not by specific reward values. 
In this section, for simplicity and clarity we use $k$-Tiered Reward defined by:
\begin{equation*}
    r_i = 
        \begin{cases}
        0 & \text{if\ } i = k \\
        \frac{1}{1-\gamma} r_{i+1} - \delta & \text{if \ } i < k \\
        \end{cases}
\end{equation*}
where $\delta$ is a small constant used to satisfy the strict inequality constraint.

Empirically, we show Tiered Reward provides faster learning on multiple tabular domains.
We further extend our results to environments with high-dimensional image inputs and deep RL algorithms.
Finally, we also explore the influence of the number of tiers and find that having more tiers can induce faster learning.

\subsection{Fast Learning with Tiered Reward}
\begin{figure}[tpb]
    \centering
    \raisebox{-0.5\height}{%
        \begin{subfigure}[b]{0.18\textwidth}
            \centering
            \includegraphics[width=\textwidth]{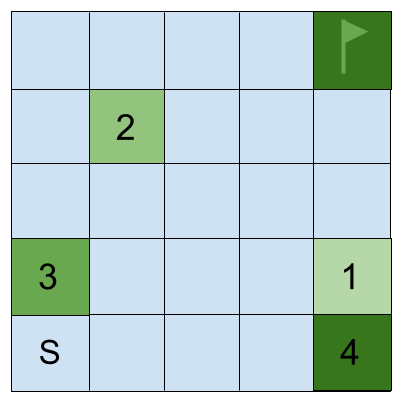}
            \caption{Flag Grid. The state space is the location of the agent plus the flag it has collected.}
            \label{fig:flag-grid}
        \end{subfigure}%
    }
    \hfill
    \raisebox{-0.5\height}{%
        \begin{subfigure}[b]{0.78\textwidth}
            \centering
            \includegraphics[width=0.49\textwidth]{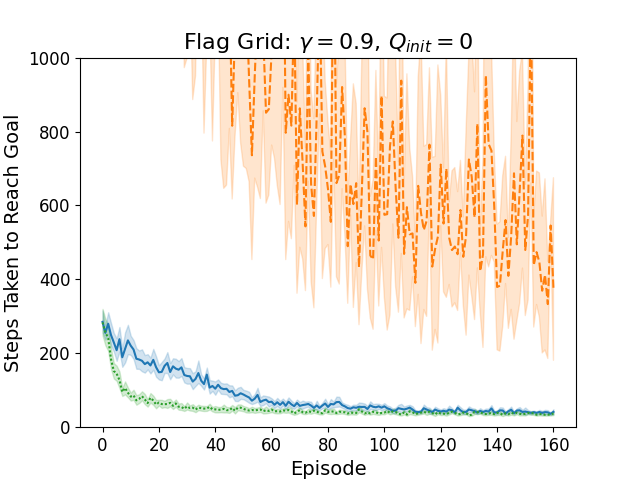}
            \includegraphics[width=0.49\textwidth]{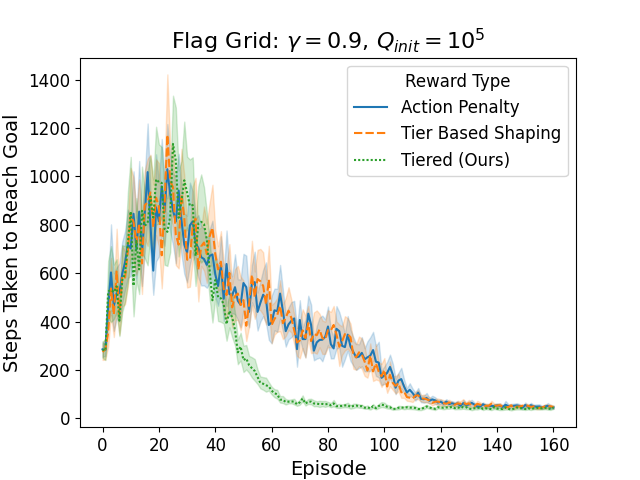}
            \caption{Q Learning curves for Flag Grid. The two plots use different initial Q-values. $\gamma \in \{0.99, 0.95, 0.90, 0.85\}$ and $Q_{init} \in \{ 10^{10}, 10^5, 10^3\}$ all had similar performances, so we only report $\gamma=0.90, Q_{init} = 10^5$ here. Error bar show standard deviation from $30$ seeds.}
            \label{fig:flag-grid-learning}
        \end{subfigure}%
    }
    \caption{Tiered Reward leads to fast learning on the Flag Grid.}
\end{figure}

The ``Flag Grid'' from \citet{ng1999policy} is a natural $6$-Tier MDP to study Tiered Reward. In this grid world (Figure~\ref{fig:flag-grid}), the agent begins in the bottom-left corner and must learn to pick up four flags in sequence before reaching the goal. 
The agent can move in $4$ directions with an $80\%$ success rate, while acting randomly $20\%$ of the time. 
All states in which the agent possesses the same number of flags constitute a tier, totaling 5 tiers. The goal constitutes the sixth and final tier.

We evaluate all three reward functions with Q-Learning~\citep{watkins1992q}, arguably the most well-understood and widely applicable RL algorithm.
As Figure~\ref{fig:flag-grid-learning} (left) shows, Tiered Reward learns fastest of the three. Perhaps surprisingly, Tier Based Shaping performs orders of magnitudes worse than even action penalty. That is because, with discounting, the shaping function $F(s, a, s') = \gamma \Phi(s') - \Phi(s)$ becomes positive when $s$ and $s'$ belong to the same tier. As a result, zero-initialized Q-values are not optimistic with respect to these rewards, and so exploration with $R_{tbs}$ is undirected and slow. For a fairer comparison, we also plot the learning curves where all Q-values are initialized to some arbitrary large value so that $R_{tbs}$ also enjoys directed exploration as the two other rewards (Figure~\ref{fig:flag-grid-learning}, right). There is a noticeable spike in time taken to reach the goal early on during training. It is the result of optimistic 
Q-value initialization, which leads to more exploration and thus slower learning. Regardless of this trade off, Tiered Reward still consistently outperforms the two baselines. 

It is important to note that our goal here is not to argue shaping is ineffective, nor to determine how to initialize Q-values for fast learning, but solely to demonstrate the usefulness of Tiered Reward in various different settings. To start, it makes learning faster than tier-based shaping reward and action penalty for different discount factors and Q-value initialization schemes. Moreover, it is simple to design and implement; there is no need to engineer environment-specific reward structures and initial Q-values to accelerate learning.

\subsection{Tiered Reward in Deep RL}

\begin{figure}[tpb]
    \centering
    \includegraphics[width=0.2\linewidth]{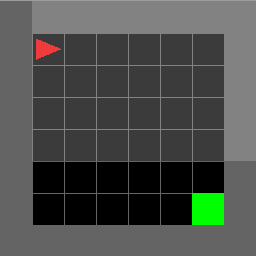}
    \includegraphics[width=0.2\linewidth]{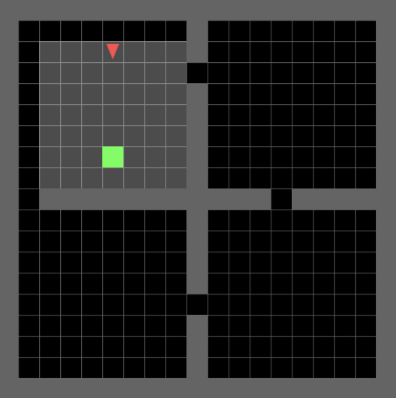}
    \includegraphics[width=0.2\linewidth]{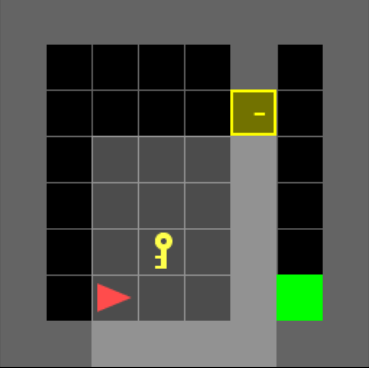}
    \caption{Illustration of EmptyGrid, FourRooms, and DoorKey from left to right. The agent is in red, and the goal in green. The objective is to navigate to the goal; in FourRooms, the agent has to find the gaps in the wall; in DoorKey, the agent must first pick up the key and unlock the yellow door. All three environments use visual observations.}
    \label{fig:minigrid-envs}
\end{figure}

\begin{figure}[t]
    \centering
    \includegraphics[width=0.32\linewidth]{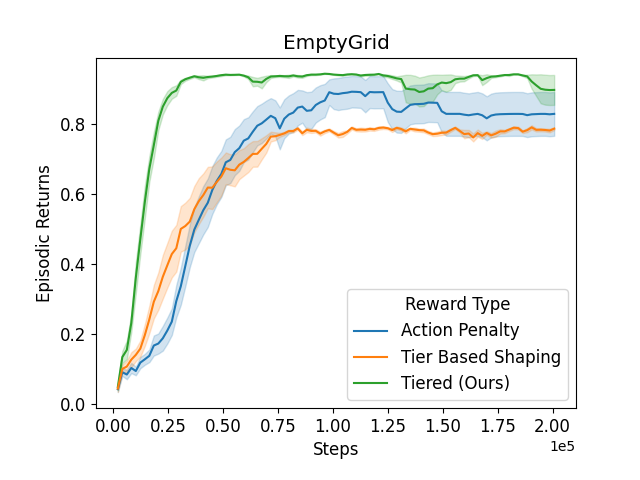}
    \includegraphics[width=0.32\linewidth]{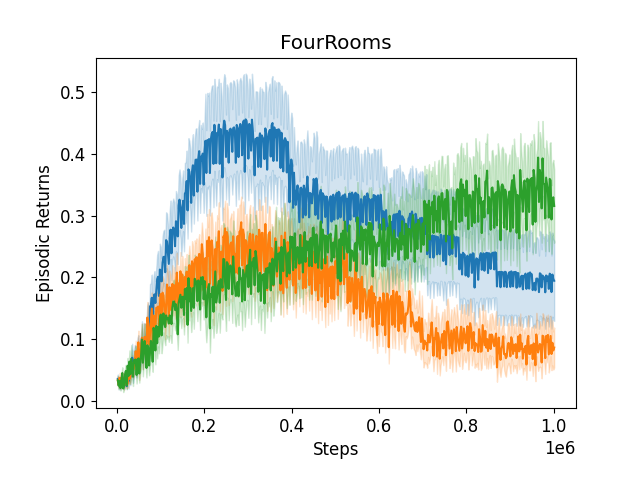}
    \includegraphics[width=0.32\linewidth]{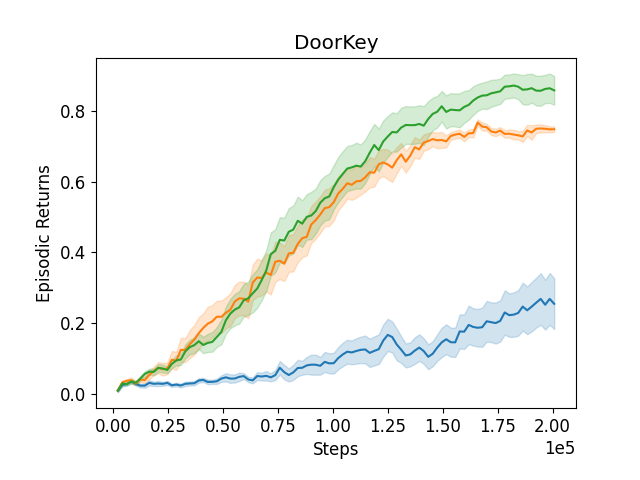}
    \caption{Learning curves of three reward functions on EmptyGrid, FourRooms, and DoorKey. Each agent is trained with the reward function labeled on the plot, but evaluated using the original MiniGrid reward ($1 - 0.9 * \text{step count} / \text{max steps}$ for success, and $0$ for failure). Error bars show standard deviation from $30$ random seeds.}
    \label{fig:minigrid-results}
\end{figure}

We further show that Tiered Reward also make learning faster in Deep RL with image observations. We choose three goal--obstacle environments from MiniGrid~\citep{Minigrid_2023}: EmptyGrid, FourRooms, and DoorKey.
In all three environments (Figure~\ref{fig:minigrid-envs}), the agent aims to learn a policy to navigate to the goal using image observations.
FourRooms is a long-horizon problem; the DoorKey environment has a complicated transition function and action space, and is hard to solve using classical RL algorithms with a sparse reward.
In all three environments, we design Tiered Reward with three tiers.
For EmptyGrid and FourRooms, tiers are assigned based on each state's $L1$ distance to the goal; for DoorKey, tiers are assigned based on the sub-goals the agent has completed (getting key, opening door, and reaching goal). 

We use Proximal Policy Optimization~\citep{schulman2017proximal} (PPO) as the RL algorithm.
To provide numerical stability, deep RL methods often employ reward scaling and clipping~\citep{henderson2018deep}.
To follow, we linearly scale the tiered reward values to be between $-1$ and $0$.
More experiment details are included in Appendix~\ref{apdx:minigrid-experiment}. 

For evaluation, we plot the episodic return with respect to the original reward function from MiniGrid~\citep{Minigrid_2023}. The original reward functions in MiniGrid are designed by human experts and express the task of reaching the goal quickly and avoiding obstacles.
Thus, performance on the original reward function shows how quickly the agent has learned the specified behavior.
Figure~\ref{fig:minigrid-results} shows the learning performance of PPO trained on different reward functions on MiniGrid. 
Across the three environments, Tiered Reward show the fastest learning, outperforming the Action Penalty and Tier Based Shaping baselines. 
Tier Based Shaping performs differently according to how hard the exploration problem is: as discussed before, when $s$ and $s'$ are in the same tier, the shaping function $F(s, a, s') = \gamma \Phi(s') - \Phi(s) > 0$, and therefore encourages the agent to exploit rather than explore.
For FourRooms, Tiered Reward initially learns slower than Tier Based Shaping, but later catches on and eventually outperforms its counterpart. 
All three reward functions suffer from large standard deviation in FourRooms likely because this environment is a hard exploration problem; different exploration during learning leads to wildly different outcomes.

\subsection{Influence of More Tiers}
\begin{figure}[tb]
    \centering
    \hfill
    \raisebox{-0.5\height}{%
        \begin{subfigure}[b]{0.18\textwidth}
            \centering
            \includegraphics[width=\textwidth]{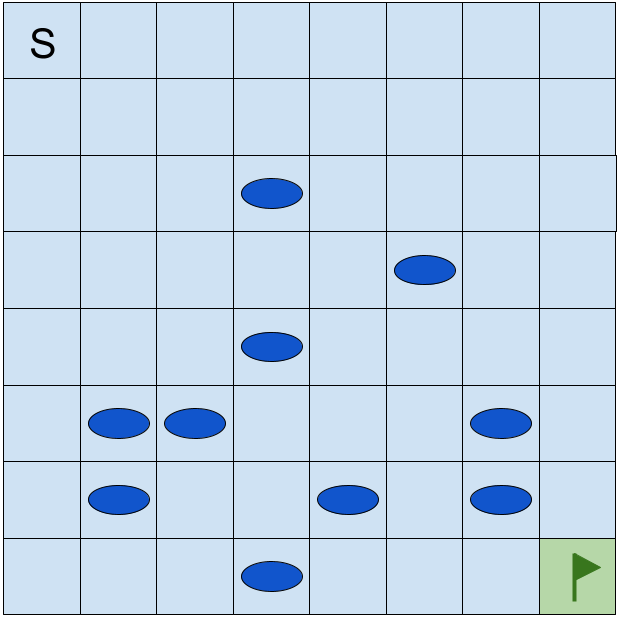} 
            \\
            \bigskip
            \includegraphics[width=\textwidth]{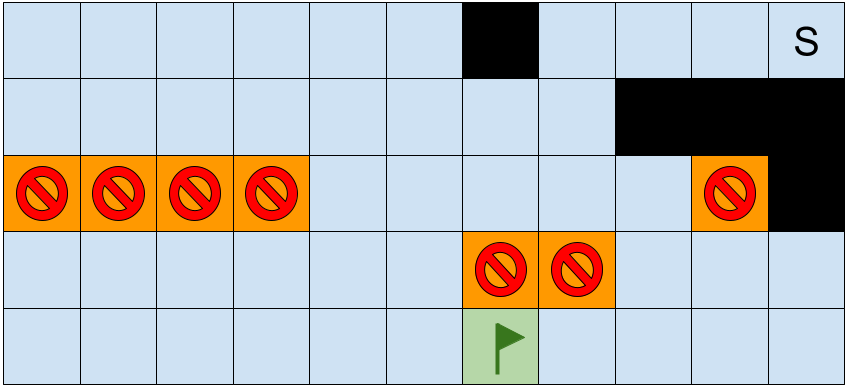}
            \\
            \caption{Top: Frozen lake environment with holes in dark blue. \\ Bottom: Wall Grid environment with walls in black and lava in orange.}
            \label{fig:frozen-lake-wall-grid}
        \end{subfigure}%
    }
    \hfill
    \raisebox{-0.5\height}{%
        \begin{subfigure}[b]{0.78\textwidth}
            \centering
            \includegraphics[width=0.44\linewidth]{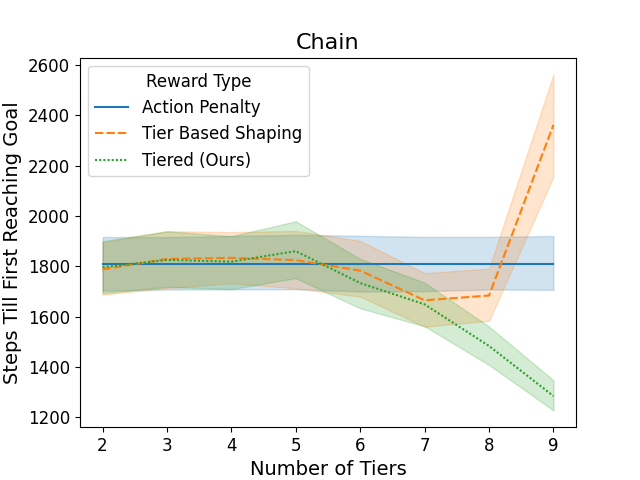}
            \includegraphics[width=0.44\linewidth]{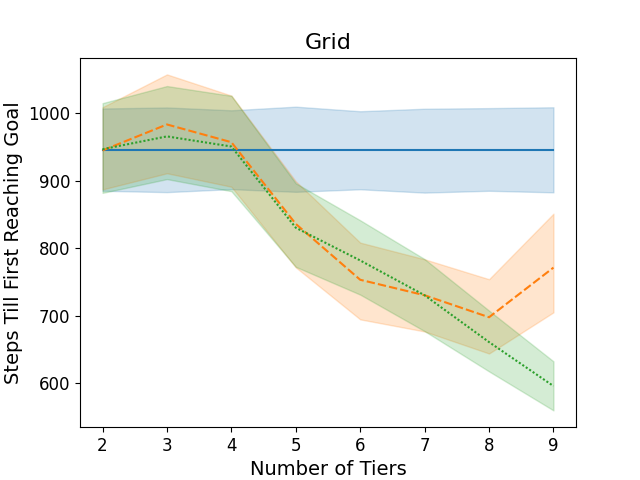}
            \includegraphics[width=0.44\linewidth]{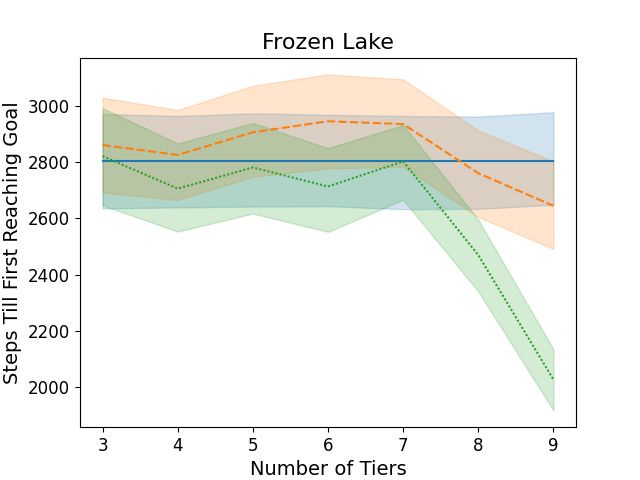}
            \includegraphics[width=0.44\linewidth]{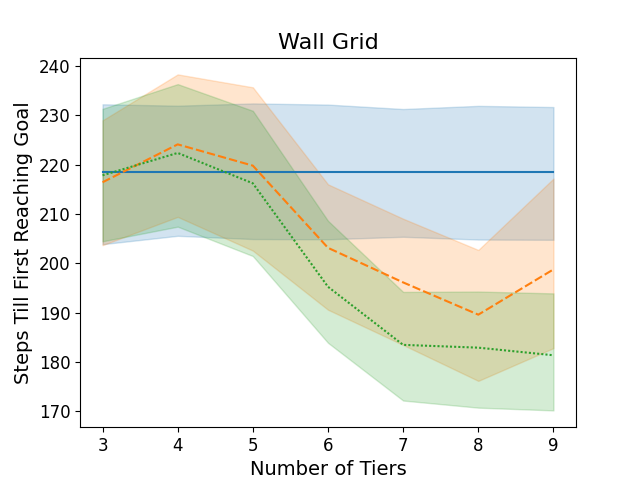}
            \caption{Q-Learning results with $\gamma=0.9$. Since there are $3$ types of states in the latter two environments, we start at $3$ tiers instead of $2$. Error bar show standard deviation from $300$ seeds.}
            \label{fig:chain-grid-learning}
        \end{subfigure}%
    }
    \hfill
    \caption{Influence of more tiers on four grid worlds with Q learning. RMAX results in Figure~\ref{fig:chain-grid-rmax}.}
\end{figure}

Finally, we explore Tiered Reward with a varying number of tiers. First, we choose for simplicity and clarity four grid-world domains with absorbing goals and obstacles that are suited to a flexible number of tiers:
\begin{enumerate}
    \item Chain: a 90--state 1D environment with left and right actions. Starting from one end, the agent tries to reach the other end with actions of success rate $80\%$; failed actions transition to the opposite direction.
    \item Grid: a $9 \times 9$ grid where the agent starts in one corner and aims for the opposite corner. The agent can move in four cardinal directions with a $80\%$ success rate, while slipping to either side with a $10\%$ chance.
    \item Frozen Lake: a slippery grid with holes that will swallow the agent (Figure~\ref{fig:frozen-lake-wall-grid} top). The objective is to reach the goal without falling into any holes. Each of the $4$ directional actions succeed $1/3$ of the time, and slip to either side with probability $1/3$.
    \item Wall Grid: a grid world with multiple lava states and wall states (Figure~\ref{fig:frozen-lake-wall-grid} bottom). The agent has to circle around the walls while avoiding the lava to get to the goal. Transition dynamics same as Grid.
\end{enumerate}
For Chain and Grid, tiers are decided based on their $L1$ distance to the goal; for Frozen Lake and Wall Grid, tiers are based on the sum of $L1$ distance to the goal and start state, weighted $2: 1$.

In absence of a ``correct" reward function that expresses the tasks, we measure the learning speed by recording the steps required for the agent to reach the goal for the first time. 
For fair comparison, we optimistically initialize $Q_{init} = 10^5$ so that $R_{tbs}$ also enjoys the benefit of directed exploration.
The results are presented in Figure~\ref{fig:chain-grid-learning}. 
We repeat the same experiments with a model-based RL algorithm, RMAX~\citep{brafman02}. The results are similar to that of Q-learning (Figure~\ref{fig:chain-grid-rmax} in Appendix~\ref{apdx:rmax-results}).
As expected, Tiered Reward makes learning faster as the number of tiers increases because more information about the environment aids reward design. Tiered Reward consistently beats action penalty and is at least as good as tier-based shaping reward, and often much better. Even when Tiered Reward performs the same as shaped reward, it provides the added benefit of simplicity and better interpretability---it is based only on states, and not $(s, a, s')$ triples.

\begin{figure}[ht]
    \centering
    \includegraphics[width=0.32\linewidth]{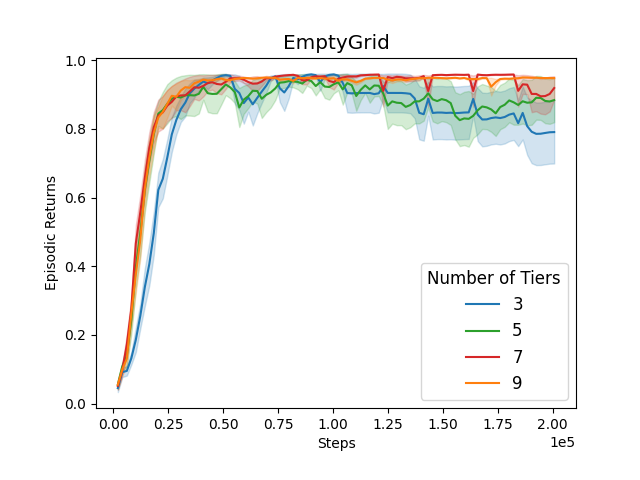}
    \includegraphics[width=0.32\linewidth]{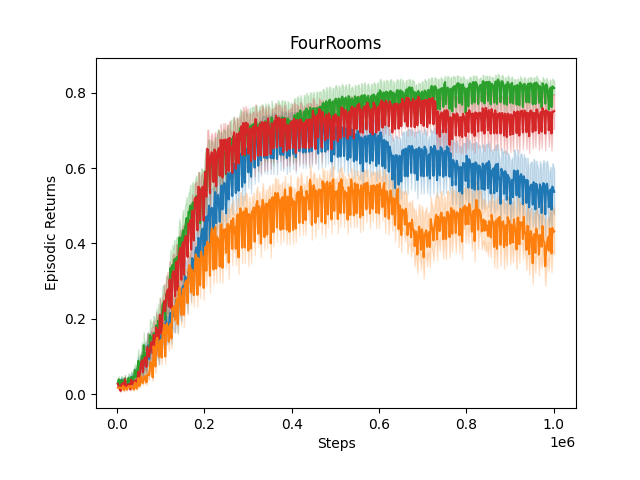}
    \includegraphics[width=0.32\linewidth]{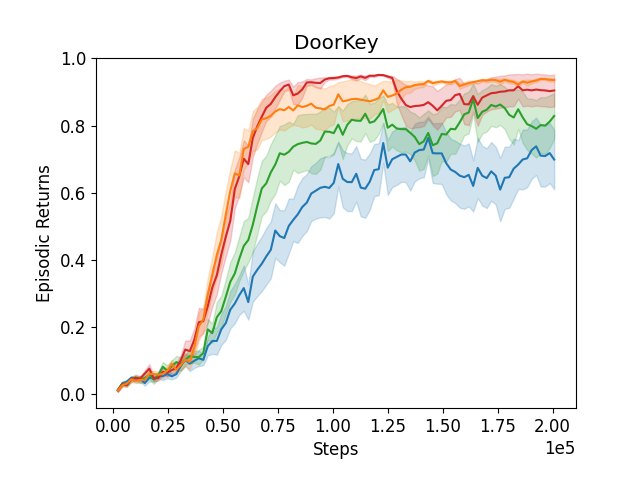}
    \caption{Learning curves on EmptyGrid, FourRooms, and DoorKey with different number of tiers in Tiered Reward. Error bars show standard deviation from $30$ random seeds.}
    \label{fig:multi-tier}
\end{figure}

Adding more tiers achieves similar effects in deep RL settings under numerical constraints. 
Figure~\ref{fig:multi-tier} shows the learning performance of Tiered Reward with $3$, $5$, $7$, and $9$ tiers on MiniGrid environments.
In EmptyGrid, having more tiers makes learning faster and more stable.
In DoorKey, $7$ and $9$ tiers produce similar results, both significantly faster than $3$ or $5$ tiers.
In FourRooms, $5$ tier leads to the fastest learning result, and using more than $5$ tiers \textit{monotonically} decreases learning performance. 
We believe this is because of numerical issues during scaling of Tiered Reward: we scale reward values to $[-1, 0]$ for stability in learning, but it also decreases the difference of reward value among tiers, especially among higher (more desirable) tiers.
Reward values for all states in tiers higher than $3$ gets normalized to very close to $0$ (in the range $10^{-6}$ to $10^{-15}$), and therefore tiers become indistinguishable under some numerical accuracy. 
Having more tiers suffers more from this problem because more states become indistinguishable, and as a result have worse performance. \footnote{To alleviate the numerical issues caused by scaling, we set discount factor to $0.5$ in these experiments to obtain smaller original Tiered Reward. See Appendix~\ref{apdx:reward-values} for further details}
We hypothesize that this issue can be resolved with a smarter reward scaling or clipping method, and leave that for future work.
In practice, the optimal number of tiers can be determined a priori with MDP-specific information or empirically.
Though more tiers sometimes leads to slower learning than $3$ tiers, all Tiered Rewards provide faster learning than the two reward baselines in Figure~\ref{fig:minigrid-results}.

\section{Conclusion}
\label{sec:conclusion}

In contrast to standard reward-design solutions that are environment-dependent, we presented Tiered Rewards---a class of environment-independent reward structures that provably leads to (Pareto) optimal behavior and empirically leads to fast learning.
Tiered Reward can be defined in both tabular and high dimensional environments, and is RL-algorithm agnostic.
Interesting future work includes getting theoretical guarantees that Tiered Reward lead to asymptotically faster learning and addressing the numerical issues with more tiers.

\subsubsection*{Acknowledgments}
\label{sec:ack}
This research is supported in part by the Office of Naval Research (ONR) award N00014-20-1-2115.

\bibliography{main}
\bibliographystyle{rlc}

\newpage
\appendix

\section{Proof of Theorem~\ref{thm:three-tier}}
\label{apdx:thm-1-proof}

\begin{proof}
Let $\pi^*$ be the optimal policy induced with Tiered Reward $R(s)$. Suppose, for the sake of contradiction, there exists some policy $\pi$ that dominates $\pi^*$. Then, by our definition of Pareto dominance, 
$$\sum_{i=0}^{t} o_i \leq \sum_{i=0}^{t} o_i^*, \quad \forall t = 0,1,2,...,\infty,$$
$$\sum_{i=0}^{t} g_i \geq \sum_{i=0}^{t} g_i^*, \quad \forall t = 0,1,2,...,\infty,$$
where $o_t$ and $g_t$ are the probabilities of reaching obstacles and goals in exactly $t$ steps following $\pi$, and $o_t^*$ and $g_t^*$ are the same for $\pi^*$.
We can write the value function (of $\pi$ being evaluated on $R(s)$) as
$$V = \sum_{t=0}^{\infty} g_t (\gamma^t r_{goal} + \sum_{j=0}^{t-1} \gamma^j r_{back}) + o_t (\gamma^t r_{obs} + \sum_{j=0}^{t-1} \gamma^j r_{back}).$$
The value of $\pi^*$ ($V^*$) can be written similarly.
Denote
$$f_t^g = \gamma^t r_{goal} + \sum_{j=0}^{t-1} \gamma^j r_{back}, \rm{\ and}$$
$$f_t^o = \gamma^t r_{obs} + \sum_{j=0}^{t-1} \gamma^j r_{back}.$$
That is, $f^t_g$ is the reward obtained on a trajectory that reaches a goal in $t$ steps and $f^t_o$ is the reward obtained on a trajectory that reaches an obstacle in $t$ steps.
With $r_{obs} < \frac{1}{1-\gamma} r_{back} < r_{goal}$, we show below that $f_t^g$ is strictly decreasing and $f_t^o$ strictly increasing with respect to $t$.

Proof that $f_t^g$ is strictly decreasing: 
\begin{equation*}
    \begin{split}
        f_{t+1}^g - f_t^g &= \gamma^{t+1} r_{goal} + \sum_{j=0}^{t}\gamma^j r_{back} - \gamma^{t} r_{goal} - \sum_{j=0}^{t-1}\gamma^j r_{back} \\
        &= \gamma^t (\gamma  - 1) r_{goal} + \gamma^t r_{back} \\
        &= \gamma^t (1 - \gamma) (\frac{1}{1-\gamma} r_{back} - r_{goal}) \\
        &< 0 
    \end{split}
\end{equation*}
because $0 < \gamma < 1$ and $\frac{1}{1 - \gamma} r_{back} < r_{goal}$.
\bigskip

Proof that $f_t^o$ is strictly increasing:
\begin{equation*}
    \begin{split}
        f_{t+1}^o - f_t^o &= \gamma^{t+1} r_{obs} + \sum_{j=0}^{t}\gamma^j r_{back} - \gamma^{t} r_{obs} - \sum_{j=0}^{t-1}\gamma^j r_{back} \\
        &= \gamma^t (\gamma  - 1) r_{obs} + \gamma^t r_{back} \\
        &= \gamma^t (1 - \gamma) (\frac{1}{1-\gamma} r_{back} - r_{obs}) \\
        &> 0 
    \end{split}
\end{equation*}
because $0 < \gamma < 1$ and $r_{obs} < \frac{1}{1 - \gamma} r_{back}$.
\bigskip

Then,
\begin{equation*}
    \begin{split}
        V - V^* &= \sum_{t=0}^\infty (g_t - g_t^*) f_t^g + \sum_{t=0}^\infty (o_t - o_t^*) f_t^o \\
               &= \sum_{t=0}^\infty (\sum_{j=0}^t g_j - g_j^*) (f_t^g-f_{t+1}^g) + \sum_{t=0}^\infty (\sum_{j=0}^t o_j - o_j^*) (f_t^o-f_{t+1}^o) \quad (*)\\
               &> 0 + 0 \\
               &= 0.
    \end{split}
\end{equation*}

The pass from the first equality to the second (*) is justified as follows:
\begin{equation*}
    \begin{split}
        \sum_{t=0}^\infty (g_t - g_t^*)f_t^g &= \sum_{t=0}^\infty \sum_{j=0}^t (g_j - g_j^*) f_t^g - \sum_{t=0}^\infty \sum_{j=0}^{t-1} (g_j - g_j^*)f_t^g  \\
        &= \sum_{t=0}^\infty \sum_{j=0}^t (g_j - g_j^*) f_t^g - \sum_{t=1}^\infty \sum_{j=0}^{t-1} (g_j - g_j^*)f_t^g \\
        & = \sum_{t=0}^\infty \sum_{j=0}^t (g_j - g_j^*) f_t^g - \sum_{t’=0}^\infty \sum_{j=0}^{t’} (g_j - g_j^*)f_{t’+1}^g \\
        & = \sum_{t=0}^\infty (\sum_{j=0}^t g_j - g_j^*) (f_t^g - f_{t+1}^g) \\
    \end{split}
\end{equation*}
Similarly, 
$$\sum_{t=0}^{\infty} (o_t - o_t^*)f_t^o = \sum_{t=0}^\infty (\sum_{j=0}^t o_j - o_j^*) (f_t^o - f_{t+1}^o)$$

We have shown, through the value function, that $\pi$ is strictly better than $\pi^*$ with respect to the reward function $R$. But $\pi^*$ was chosen to optimize $R$, so that's a contradiction. 
Since no such $\pi$ can exist, that means   
$\pi^*$ is not dominated by any policy, and is therefore Pareto-optimal.
\end{proof}

\newpage
\section{Proof of Theorem~\ref{thm:multi-tier}}
\label{apdx:thm-multi-tier-proof}
\begin{proof}
The proof is similar to that of Theorem~\ref{thm:three-tier}.
Suppose, for the sake of contradiction, that there exists some such policy $\pi$. We can express the value functions as
$$V = \sum_{t=0}^{\infty} \gamma^t \sum_{m=1}^{k} r_m \cdot p_t^m, {\rm \ and}$$
$$V^* = \sum_{t=0}^{\infty} \gamma^t \sum_{m=1}^{k} r_m \cdot {p_t^*}^m.$$
Denote 
$f_t^m = \gamma^t r_m $.
Then,
$f_t^m - f_{t+1}^m = r_m \gamma^t (1-\gamma) \leq 0, \forall m$. It's easy to see $f_t^m - f_{t+1}^m$ is strictly increasing in $m$, so
\begin{equation*}
    \begin{split}
        V - V^* &= \sum_{t=0}^{\infty} \gamma^t \sum_{m=1}^{k} r_m (p_t^m - {p_t^*}^m) \\
                &= \sum_{m=1}^{k} \sum_{t=0}^{\infty} f_t^m (p_t^m - {p_t^*}^m) \\
                &= \sum_{m=1}^{k} \sum_{t=0}^{\infty} (f_t^m - f_{t+1}^m) (\sum_{j=0}^{t}p_j^m - {p_j^*}^m) \\
                &> \sum_{m=1}^{k} \sum_{t=0}^{\infty} (f_t^1 - f_{t+1}^1) (\sum_{j=0}^{t}p_j^m - {p_j^*}^m) \quad (**)\\
                &= \sum_{t=0}^{\infty} (f_t^1 - f_{t+1}^1) \sum_{m=1}^{k} (\sum_{j=0}^{t}p_j^m - {p_j^*}^m) \\
                &= \sum_{t=0}^{\infty} (f_t^1 - f_{t+1}^1) \cdot \sum_{j=0}^{t} (\sum_{m=1}^{k}p_j^m - \sum_{m=1}^{k}{p_j^*}^m) \\
                &= \sum_{t=0}^{\infty} (f_t^1 - f_{t+1}^1) \cdot \sum_{j=0}^{t} (1-1) \\
                &= 0
    \end{split}
\end{equation*}
Note that the $(**)$ step is justified only because $\sum_{j=0}^{t}p_j^m - {p_j^*}^m \geq 0, \forall m = [2..k], \forall t$.
The inequalities show that $\pi$ achieves higher reward than the optimal policy, which is a contradiction. No such $\pi$ exists.
\end{proof}

\newpage
\section{Visualization of Pareto-optimal Policies}
\label{apdx:pareto}

Going back to the example of the Russell/Norvig grid, we can visualize how the probability of reaching the goal ($g_t$) and reaching lava ($o_t$) changes over time for different policies. Consider two simple policies on the Russell/Norvig grid---(1) going left from all states (``always left'') and (2) going right from all states (``always right'').

\begin{figure}[!htbp]
    \centering
    \includegraphics[width=0.5\linewidth]{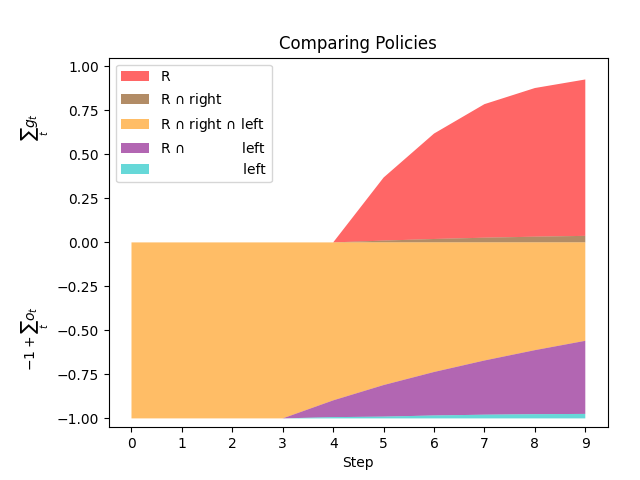}
    \caption{Visualization of the policies of always going left and always going right in the (stochastic) Russell/Norvig grid. The policy $R$ is the same as $R$ in Figure~\ref{fig:rgb-policy}. To avoid color overlapping, we separated each policy into disjoint regions visualized by distinct colors. Each colored region in the figure represent the probability-region of one or more policies, joined by $\cap$. For example, the policy ``always right'' covers the areas in brown and orange.}
    \label{fig:rn-one-direction-policy}
\end{figure}
We visualize each policy as a shaded area upper bounded by $\sum_t g_t$ and lower bounded by $-1 + \sum_t o_t$ in Figure~\ref{fig:rn-one-direction-policy}. This visualization can be understood as separating the probability space into two, with the goal-reaching probability on the top half of the y-axis in $[0, 1]$ and obstacle-hitting probability in the bottom half of the y-axis in $[-1, 0]$. With this visualization, a Pareto-dominated policy will cover an area that is entirely enclosed by that of a dominating policy because of lower goal-reaching probabilities on the top half and higher obstacle-hitting probabilities on the bottom half. As Figure~\ref{fig:rn-one-direction-policy} shows, ``always right" and ``always left" do not cover each other, so they are incomparable. Specifically, ``always right" has a slightly higher probability of reaching the goal (brown), but ``always left' has a lower probability of reaching the lava (purple and teal). 

For comparison, we plot another policy, which we call $R$, that is state-dependent and moves in the direction of the goal. For policy $R$, the probability of reaching the target increases with time because each step has a 20\% slip probability; agents could slip early on and take longer to reach the goal. Note that area covered by $R$ (red, brown, orange, and purple) completely subsumes that of ``always right'' (brown and orange), demonstrating that ``always right" is dominated by $R$. ``Always left'', on the other hand, is not dominated by $R$ because it has a lower probability of reaching lava (teal). However, ``always left'' is not Pareto-optimal of course, because it is dominated by policy $G$ in Table~\ref{tab:three-reward-functions}. 

\begin{figure}[!htpb]
    \centering
    \includegraphics[width=0.5\linewidth]{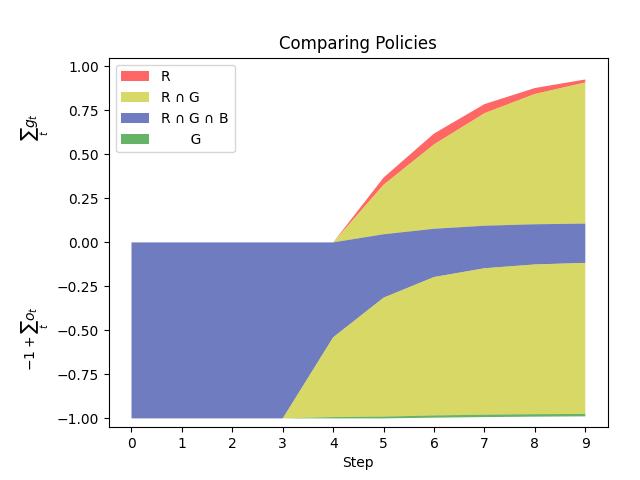}
    \caption{Visualization of three different policies (R, G, B) on Russell/Norvig grid. Visualization scheme is the same as described in Figure~\ref{fig:rn-one-direction-policy}.}
    \label{fig:rgb-policy}
\end{figure}

In Figure~\ref{fig:rgb-policy}, we visualize the three policies $R$, $G$, and $B$ in Table~\ref{tab:three-reward-functions}. Both $R$ and $G$ are Pareto-optimal, while $B$ is Pareto-dominated because $B$'s areas are entirely enclosed by that of $R$ and of $G$.

\clearpage
\section{Tabular Grid Worlds Experiment Details}
\label{apdx:tabular-experiment}
The implementation of many tabular environments and algorithms is based on the MSDM library by~\citet{Ho_Models_of_Sequential_2021}. All MSDM experiments are run on an Ubuntu $18.04$ system with an Intel Core i7-9700K CPU and $32$G of RAM.

Following \citet{koenig1993complexity}, we use a greedy policy for action selection and initialize the Q-values optimistically to exploit directed exploration. Since all reward values are non-positive, it is sufficient to initialize the Q-values to $0$. We use a learning rate of $\alpha=0.90$ (tuned from $\alpha \in \{0.95, 0.90, 0.85\}$, all of which performed similarly). We set the small constant $\delta=0.1$.

\section{RMAX results}
\label{apdx:rmax-results}

We set maximal reward $r_{max} = 10^5$ , use the first $m=3$ transition samples to model the MDP (tuned from $m \in \{2, 3, 5, 7, 10\}$ and selected based on good resulting policy while taking similar learning time to Q-Learning), and did $200$ iterations of value iteration during each update. 

\begin{figure*}[htpb]
    \centering
    \includegraphics[width=0.40\linewidth]{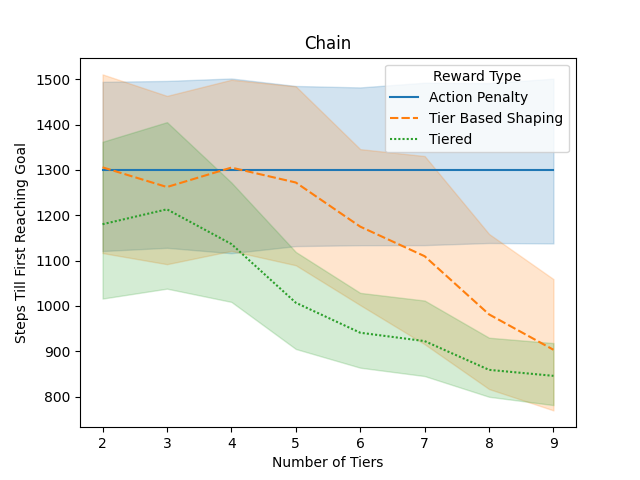}
    \includegraphics[width=0.40\linewidth]{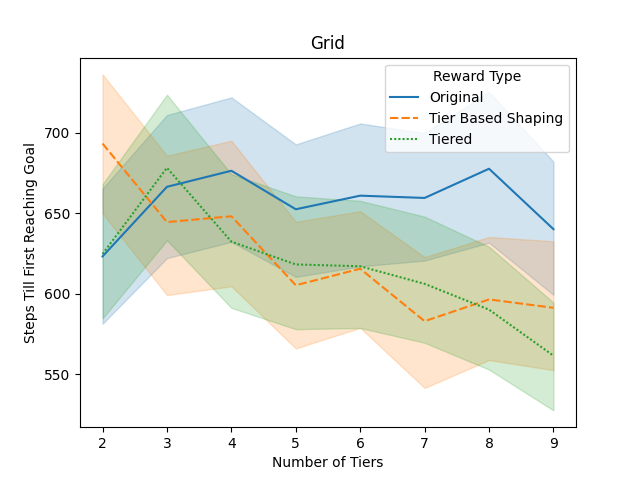}
    \includegraphics[width=0.40\linewidth]{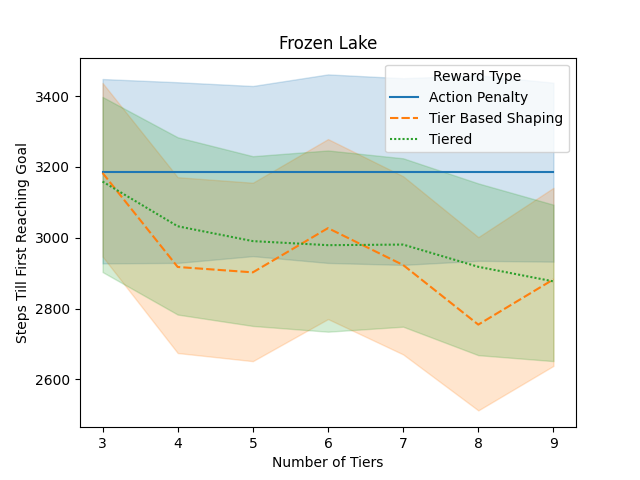}
    \includegraphics[width=0.40\linewidth]{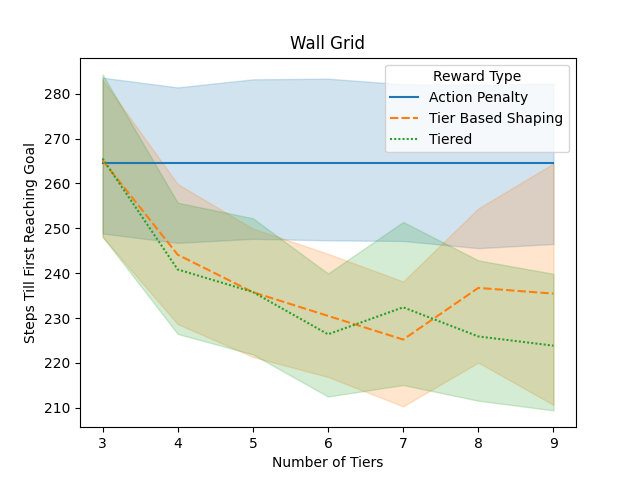}
    \caption{RMAX results with $\gamma=0.9$. Each experiment was run with $300$ seeds.}
    \label{fig:chain-grid-rmax}
\end{figure*}

\newpage
\section{MiniGrid Experiment Details}
\label{apdx:minigrid-experiment}
The DeepRL experiments on MiniGrid environment were run on an Ubuntu $18.04$ system using a single 6GB GPU (NVIDIA GeForce GTX 980 Ti) and took between 40-60min per seed, depending on the environment. 

A Proximal Policy Optimization (PPO) agent was trained on the three MiniGrid environments using the following hyperparameters:

\begin{table}[th]
\begin{center}
\begin{tabular}{lcc}
\multicolumn{1}{c}{\bf Hyperparameter}  
&\multicolumn{1}{c}{\bf Paramter Value}
&\multicolumn{1}{c}{\bf Note}
\\ \hline \\
        Epochs & 4 & \\
        Batch Size & 256 & \\
        Learning Rate ($\alpha$) &  $1\times10^{-3}$\\
        Total Env Steps & 200000 & Total training steps in each environment\\
        Discount Factor ($\gamma$) & 0.5 & See Appendix~\ref{apdx:reward-values} \\
        Small Constant ($\delta$) & 5 & Used to satisfy the strict inequality in Tiered Reward \\
        GAE $\lambda$ & 0.95 & $\lambda$ coefficient in the GAE formula \\
        Entropy Coefficient & 0.01 & \\
        Value Loss Coefficient & 0.5 & \\
        Max Grad Norm & 0.5 & Maximum norm of gradient \\
        Clipping $\epsilon$ & 0.2 & \\
        
\end{tabular}
\end{center}
\caption{PPO hyperparameters on MiniGrid experiments.}
\label{tab:minigrid-ppo-hyperparams}
\end{table}

The PPO agent used an ImpalaCNN architecture~\citep{espeholt2018impala} for the policy and value network. The architecture of the policy network is as follows:

\begin{table}[th]
\begin{center}
\begin{tabular}{ccc}
\multicolumn{1}{c}{\bf Layer Name}  &\multicolumn{1}{c}{\bf Layer Type}
&\multicolumn{1}{c}{\bf Output Dimension}
\\ \hline \\
        Conv Sequence 1 & Convolutional Layers & 16 \\
        Conv Sequence 2 & Convolutional Layers & 32 \\
        Conv Sequence 3 & Convolutional Layers & 32 \\
        Hidden Layer & Linear Layer & 256 \\
        Logit Layer & Linear Layer & 256 \\
        Value Layer & Linear Layer & 256 \\
\end{tabular}
\end{center}
\caption{Architecture of the policy network of the Proximal Policy Optimization agent used in MiniGrid experiments.}
\label{tab:impala-arch}
\end{table}

Each Conv Sequence in Table~\ref{tab:impala-arch} is a sequence of convolutional and pooling layers with residual connections. 
Each convolutional and pooling layer in the conv sequence has the same number of output channels, as specified in the output dimension in Table~\ref{tab:impala-arch}.
The architecture of a Conv Sequence is as follows:
\begin{table}[th]
\begin{center}
\begin{tabular}{ccccc}
\multicolumn{1}{c}{\bf Layer \#}  
&\multicolumn{1}{c}{\bf Layer Type}
&\multicolumn{1}{c}{\bf Kernel Size}
&\multicolumn{1}{c}{\bf Stride}
&\multicolumn{1}{c}{\bf Padding}
\\ \hline \\
        1 & Conv & 3 & 1 & 1\\
        2 & Max Pool & 3 & 2 & 1\\
        3 & Relu &  &  &  \\
        4 & Conv & 3 & 1 & 1 \\
        5 & Relu & & & \\
        6 & Conv & 3 & 1 & 1 \\
        7 & Relu &  &  &  \\
        8 & Conv & 3 & 1 & 1 \\
        9 & Relu & & & \\
        10 & Conv & 3 & 1 & 1 \\
\end{tabular}
\end{center}
\caption{Architecture of a Conv Sequence.}
\label{tab:impala-arch}
\end{table}

There are two residual connections: one between layers $3$ and $6$, and another between layers $7$ and $10$.

\clearpage
\newpage
\section{Visualizing Additional Tiers on Empty Grid}\label{apdx:multi-tier-visualization}

To aid the understanding of Tiered Reward, we provide below visualizations of Tiered Reward on a grid world with different number of Tiers. In this grid world, the agent starts at the bottom right corner, and the goal is the top left corner. We assign the tiers based on a state's $L_1$ distance to the goal. 

\begin{figure*}[htpb]
    \centering
    \includegraphics[width=0.40\linewidth]{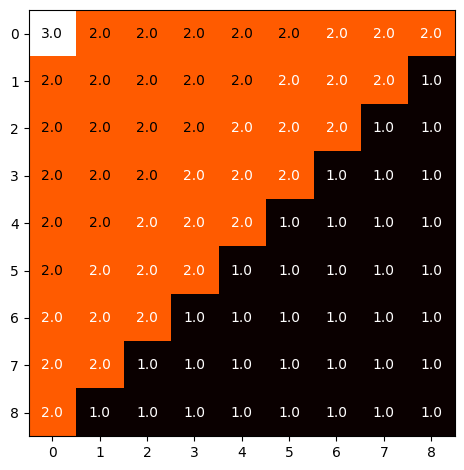}
    \includegraphics[width=0.40\linewidth]{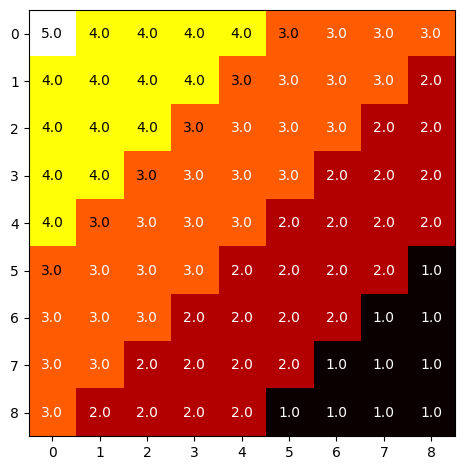}
    \includegraphics[width=0.40\linewidth]{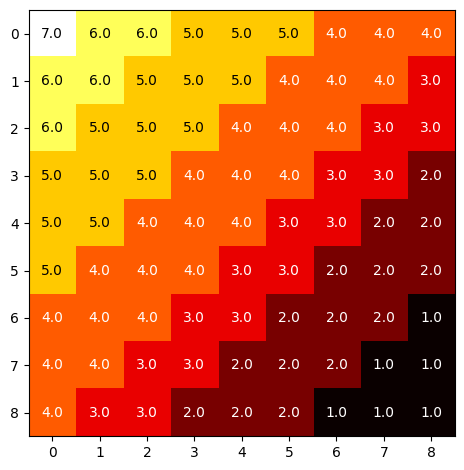}
    \includegraphics[width=0.40\linewidth]{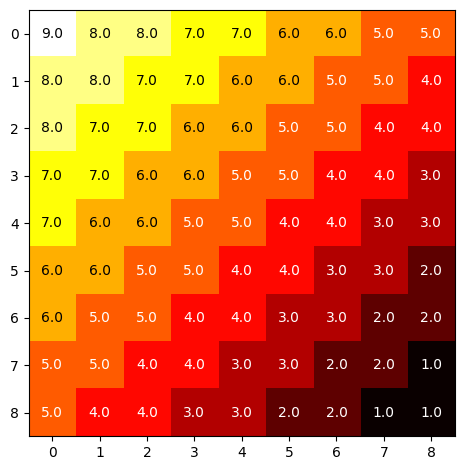}
    \caption{A visualization of Tiered Reward for $3$ (top left), $5$ (top right), $7$ (bottom left), and $9$ (bottom right) tiers on a grid world. Each color in the plot represents one single tiers, and the numbers represent the tier number.}
    \label{fig:multi-tier-visualization}
\end{figure*}

\newpage
\section{Scaled Tiered Reward for Different Tiers}
\label{apdx:reward-values}

We provide here the numerical values of the Tiered Reward after scaling it to be between $[-1, 0]$. This provides intuition on how using smaller discount values can reduce the numerical problems during reward scaling because big values of $\gamma$ (for example, $\gamma=0.99$ or $\gamma=0.9$) will make scaled reward values numerically equal for different tiers, especially the higher tiers.

\begin{table}[!h]
\begin{center}
\begin{tabular}{|c || c | l | l |} 
 \hline
 & Tier & $\gamma=0.99$ & $\gamma=0.5$ \\
\hline
\multirow{3}{4em}{Tiered Reward} & 5 & $0$ & $0$ \\ 
& 4 & $-5$ & $-5$ \\ 
& 3 & $-505$ & $-15$ \\ 
& 2 & $-50505$ & $-35$ \\ 
& 1 & $-5050505$ & $-75$  \\ 
\hline
\multirow{3}{4em}{Scaled Tiered Reward} & 5 & $0$ & $0$ \\ 
& 4 & $-9.9\times10^{-7}$ & $-0.06$ \\ 
& 3 & $-9.9\times10^{-5}$ & $-0.2$ \\ 
& 2 & $-9\times10^{-3}$ & $-0.4667$ \\ 
& 1 &  $-1$ & $-1$ \\ 
\hline
\end{tabular}
\caption{Comparing scaled and unscaled reward values using a total of $5$ tiers and $delta=5$. Reward values for two discount factors ($\gamma$) are provided.}
\label{tab:5-tiers-rewards}
\end{center}
\end{table}

\begin{table}[!h]
\begin{center}
\begin{tabular}{|c || c | l | l |} 
 \hline
 & tier & $\gamma=0.99$ & $\gamma=0.5$ \\
\hline
\multirow{3}{4em}{Tiered Reward} & 9 & $0$ & $0$ \\ 
& 8 & $-5$ & $-5$ \\ 
& 7 & $-505$ & $-15$ \\ 
& 6 & $-50505$ & $-35$ \\ 
& 5 & $-5050505$ & $-75$  \\ 
& 4 & $-505050505$ & $-155$  \\ 
& 3 & $-50505050505$ & $-315$  \\ 
& 2 & $-5050505050505$ & $-635$  \\ 
& 1 & $-505050505050505$ & $-1275$  \\ 
\hline
\multirow{3}{4em}{Scaled Tiered Reward} & 9 & $0$ & $0$ \\ 
& 8 & $-9\times10^{-15}$ & $-0.0039$\\ 
& 7 & $-9\times10^{-13}$ & $-0.0118$ \\ 
& 6 & $-9\times10^{-11}$ & $-0.2258$ \\ 
& 5 &  $-9\times10^{-9}$ & $-0.0588$ \\ 
& 4 & $-1\times10^{-6}$ & $-0.1216$  \\ 
& 3 & $-9\times10^{-5}$& $-0.2471$  \\ 
& 2 & $-0.01$ & $-0.4980$  \\ 
& 1 & $-1$ & $-1$  \\ 
\hline
\end{tabular}
\caption{Comparing scaled and unscaled reward values using a total of $9$ tiers and $delta=5$. Reward values for two discount factors ($\gamma$) are provided.}
\label{tab:9-tiers-rewards}
\end{center}
\end{table}

\newpage
\section{Additional Figure: Pareto-optimal Rewards}
\label{apdx:additional-figures}

\begin{figure}[htpb]
    \centering
    \includegraphics[width=0.6\linewidth]{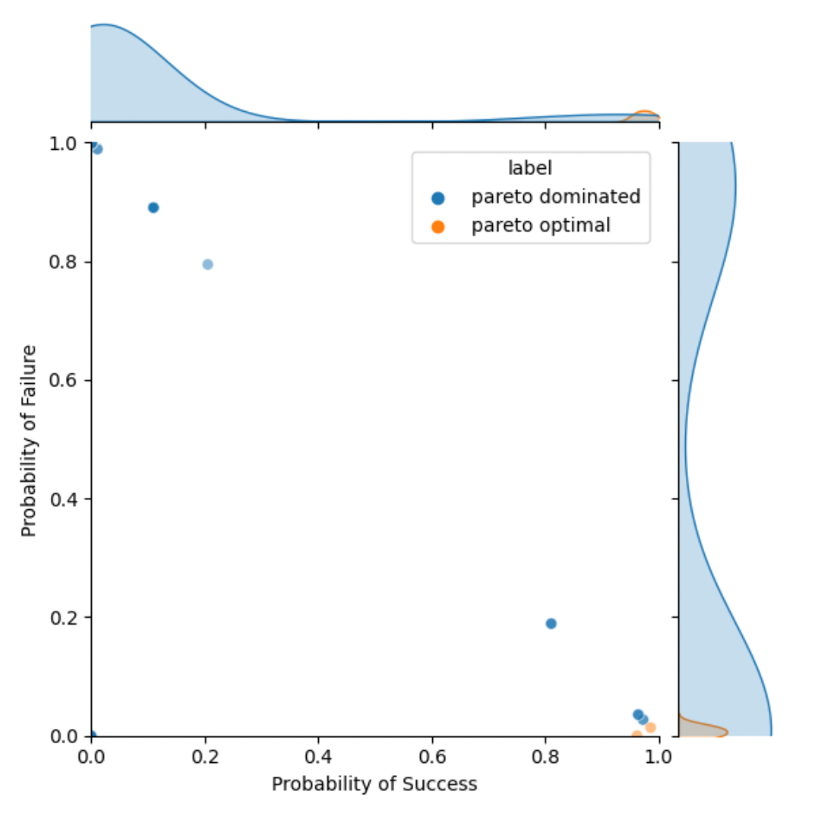}
    \caption{1000 policies that are induced by random reward functions that satisfy $r_{lava} < r_{back} < r_{goal}$ from the Russell/Norvig grid. Each point in the scatter plot represents one policy's probability of success (reaching the goal) and failure (reaching the lava), showing that the majority ($90.5\%$) of policies in the policy space are Pareto-dominated.}
    \label{fig:pareto-random-rewards}
\end{figure}

\end{document}